# Automated Medical Device Display Reading Using Deep Learning Object Detection

Lucas P. Moreira, *Federal Institute of Education, Science and Technology of Brasilia, Brasilia, Brazil*
lucas.moreira@ifb.edu.br

*Abstract*—Telemedicine and mobile health applications, especially during the quarantine imposed by the covid-19 pandemic, led to an increase on the need of transferring health monitor readings from patients to specialists. Considering that most home medical devices use seven-segment displays, an automatic display reading algorithm should provide a more reliable tool for remote health care. This work proposes an end-to-end method for detection and reading seven-segment displays from medical devices based on deep learning object detection models. Two state of the art model families, EfficientDet and EfficientDet-lite, previously trained with the MS-COCO dataset, were fine-tuned on a dataset comprised by medical devices photos taken with mobile digital cameras, to simulate real case applications. Evaluation of the trained model show high efficiency, where all models achieved more than 98% of detection precision and more than 98% classification accuracy, with model EfficientDet-lite1 showing 100% detection precision and 100% correct digit classification for a test set of 104 images and 438 digits.

## I. Introduction

THE use of digital and electronic technologies, known as e-health, is seen as one of the approaches that could help to improve the quality of health care services and, as a result, the experience of patients and other service receivers [1]. Among the possible e-health services the increasingly popular between healthcare professionals and general public are those based on telemedicine [2] and mobile health, or mHealth.

Affordable and easy to use home medical devices, especially glucometers and blood-pressure monitors, added to the extensible use of smartphones, turned possible the remote monitoring of glycaemia control on patients with diabetes [3,4,5] and blood pressure for patients with hypertension [6,7,8].

According to Finnegan et al [9], more than 70% of recommended home blood glucose meters and more than 90% of blood pressure monitors use seven-segment displays to show their measurements and less than 5% of both types of devices send the information to a smartphone via Bluetooth. The implementation of telemedicine or mHealth applications thus require the manually insertion or transmission of the device's readings.

This paper proposes an automatic seven-segment digit detection system capable of correctly reading home glucometers and blood-pressure monitor measurements to make possible a mHealth and telemedicine software for mobile devices. The proposed system is based on lightweight deep learning models with an end-to-end approach, requiring minimal pre and post data processing.

## II. Related Work

### A. 7-segment Display

Several works have addressed the seven-segment display reading for different applications, most of them using similar approach: locate the display region, localize the digits, and then classify the digits. The most common methods for the display location step are using classical image processing techniques, such as binarization, erosion-dilation, filters, etc, followed by digits recognition using either typical optical character recognition [10] or artificial neural networks [9,11,12].

These multiple steps processing usually require extra software libraries for the pre and post processing, increasing the algorithm size and processing time.

### B. Deep Learning Object Detection

The use of neural networks, in particular Convolutional Neural Networks (CNN), for object detection in 3-channel (RGB) images has been one of most advances in computer vision and with major improvements over the last decade. Since the development of one-stage detectors, or single-stage detectors [13,14], substantial advancements were observed in face recognition [15], autonomous driving [16], video object detection [17], and so on.

Several deep learning models for object detection were optimized by TensorFlow to work in embedded and mobile devices, the so called TensorFlow Lite. The model's small size and fast inference for TF-lite models make them suitable for a range of applications, such as internet-of-things [18], wearable technology [19] and human activity recognition on smartphones [20].

### C. EfficientDet

Recently, the Google Brain Team has released a novel object detection model family, the EfficientDet-D0 to D7 [21], using as backbone the image classification convolutional network EfficientNet-B0 to B7 [22]. This model has achieved state-of-the-art performance for generic object detection and, for this reason, has been successfully applied in different applications, for example target detection [23], fabric defect detection [24], diabetic foot ulcers detection [25], waste detection [26], etc.

Aiming the mobile and embedded applications, TensorFlow Lite provides a family of object detection models, EfficientDet-lite0 to lite4, using the same CNN backbones, already formatted for small model size and fast inference, with performance slightly inferior to the original family counterpart EfficientDet-D0 to D4. All available architectures can be downloaded with weights pre-trained with large datasets, such as MS-COCO [27], for transfer learning.

### III. DATASET

The dataset used in this work were made public by [9], consisting in 487 images of four different medical devices, and the images taken using different mobile cameras in real case scenarios, changing device orientation, distance to the camera, focus and amount of ambient light. Examples of this images can be seen in Figure 1. 80% of the dataset was randomly selected for training and the remaining 20% for evaluation. To avoid data leakage, each device subset was splitted separately.

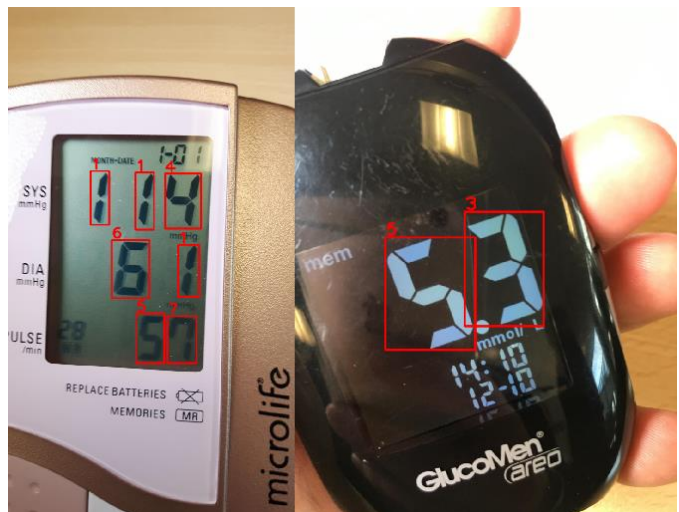

Fig. 2. Examples of bounding boxes annotation on each digit showing their boundaries and classes.

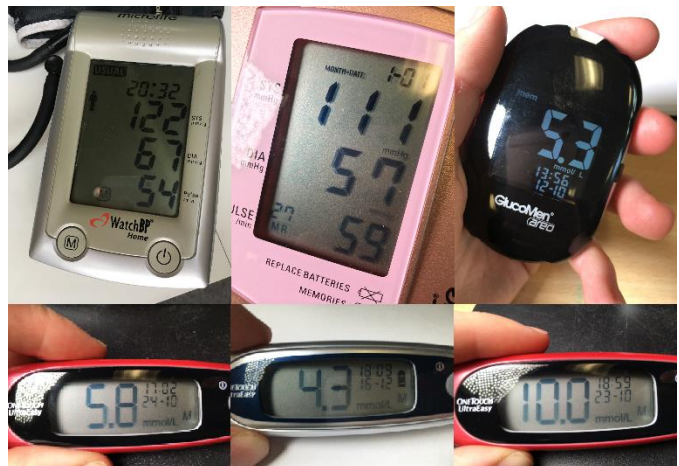

Fig. 1. Examples of images of different medical devices used for models training and evaluation.

The total number of seven-segment digits present in the dataset is 2,190, which 1719 are in the training dataset and 438 are in testing images. The bounding boxes of each digit of interest (DOI) to be detected and classified by the model were manually annotated by the author, while digits such as date and time, device's status, etc, were not annotated. As the latter are usually much smaller than the DOIs, they are not expected to interfere on model's training. Figure 2 shows examples of labelled bounding boxes.

The digits occurrences are slightly imbalanced with the digit 1 (one) with highest occurrence, 441, and the digit 0 (zero) with the lowest, 114. The distribution for all digits is shown in Figure 3. Due to variation on devices orientations related to the camera, there is a range of different aspect ratios for digits bounding boxes, additionally, the digit 1 (one) has a lower aspect ratio comparing to other digits.

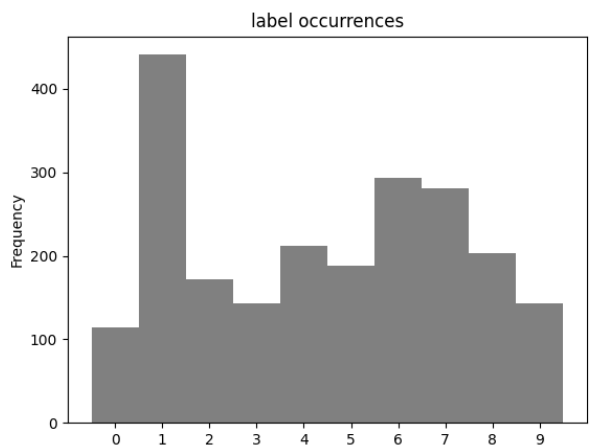

Fig. 3. Histogram of each label in the dataset.

### IV. METHODS

This section presents the proposed single-stage detection framework for seven-digit detection and classification for both architecture families, the EfficientDet and EfficientDet-lite, as well as the hyper-parameters used in model training. All stages of models training, and evaluation were developed using the TensorFlow API for object detection.

#### A. Data Augmentation

The neural network training efficiency is highly dependent on data volume and more robust when noise is present in part of the training dataset [28]. For this reason, it was used data augmentation techniques to increase the training dataset and avoid model overfitting. The methods used for the EfficientDet family were random jitter of bounding boxes corner's locations by 5%; random adjustment of images contrast between 0.8 to 1.25 of original images; 20% random variation of images brightness; and application of JPEG with a random compression quality factor between 80% and 100% of the original image.

The Tensorflow API for the EfficientDet-lite family have less built-in image augmentation options. In order to simplify the model training framework, only bounding boxes jitter were applied.

*B. Transfer Learning*

Transfer learning is an important tool in machine learning to solve the basic problem of insufficient training data. The idea is to train a neural network on large datasets and use its trained weights as initial weights in an identical or similar model for a different target domain. This will lead to a great positive effect on many domains that are difficult to improve because of insufficient training data [29]. The most popular transfer learning source domains in computer vision tasks are the ImageNet for image classification [30] and MS-COCO for image object detection [27].

For the seven-segment object detection task, all models were trained starting their weights with pre-trained values from MS-COCO dataset. These weights are usually provided in most deep learning frameworks, including the TensorFlow API used in this work. Once the weights values are assigned to the model, they are fine-tuned using the above described seven-segment dataset.

*C. Model Architecture*

The neural network architectures chosen are of the family EfficientDet (D0, D1 and D2) including the mobile versions EfficientDet-lite (lite0, lite1 and lite2). Considering the similarities between all models used, most of their architecture hyperparameters are unchanged comparing to the models used for transfer learning, including the input tensor shape. Other hyperparameters were adjusted to fit the target domain and the dataset used for training.

The first change is the number of output classes. The MS-COCO dataset has boxes labelled with 91 different classes for general application. For the seven-segment task, only 10 classes are necessary (digits from 0 to 9), thus, all models used in this work have only 10 different output classes.

The values of aspect ratios of the bounding boxes are important hyperparameters of model training, thus the distribution of its occurrences is relevant. Common values for bounding boxes aspect ratios are 0.5, 1.0 and 2.0, which sufficiently covers objects in general cases. For the seven-segment detector, a more diverse values of aspect ratios are observed, Figure 4 shows the histogram for bounding boxes aspect ratios for the hole dataset. As it is possible to see, most digits have bounding boxes with aspect ratio between 0.8 and 1.0, which this last value is sufficient to cover these boxes. Some boxes have aspect ratio between 0.3 and 0.4 and few of them have aspect ratio less than 0.3. For this reason, bounding boxes with aspect ratio of 0.3 and 0.1 were added to all models' anchor generator.

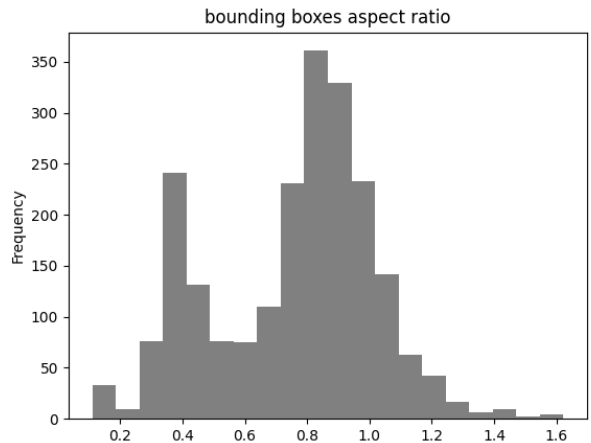

Fig. 4. Histogram of bounding boxes aspect ratio in the dataset.

*D. Pre-processing*

The proposed framework is intended to be an end-to-end object detector, requiring minor data processing before and after the model detection and classification. However, two data processing are necessary to be performed before presenting the images to the model, either for training or inference.

The first one is to rescale the image to the size expected by the model. All model architectures use square images as input tensors, thus the images are padded with zeros in order to avoid stretching effects on the DOIs. As stated above, different architectures have different input shapes, so the rescale process is not done on the dataset, it is applied for each model during the training and inference. The input image shape is summarized in Table 1.

TABLE I
INPUT TENSOR SHAPE FOR OBJECT DETECTION MODEL ARCHITECTURES

| Model Architecture | width x height (pixels) |
|---|---|
| *EfficientDet-D0* | 512 x 512 |
| *EfficientDet-D1* | 640 x 640 |
| *EfficientDet-D2* | 768 x 768 |
| *EfficientDet-lite0* | 320 x 320 |
| *EfficientDet-lite1* | 384 x 384 |
| *EfficientDet-lite2* | 448 x 448 |

The second pre-process stage is pixel value normalization by the mean and standard deviation of all image's pixels.

*E. Training*

Model training consists in presenting the labelled images to the model and perform the backpropagation weights adjustments. Due to memory limitations on the GPU, the images are presented in mini-batches, where the weights adjustments using one mini-batch is consider one training step. The training algorithm is the stochastic gradient descent (SGD) with momentum of 0.9. For the EfficientDet family, the learning rate has a base value of 0.08 with cosine decay, and trained for 30,000 steps, while for the EfficientDet-lite family the learning rate is fixed in 0.08 and the model is trained for 100 epochs, where one epoch is considered finished when all training samples are presented to the model.

*F. Post-processing*

During training and inference, the model's output for an input image is a list of proposed bounding boxes with their location (corner coordinates), the classification class and the classification confidence, also called classification score (from 0.0 to 1.0). Usually, a high number of bounding boxes are generated with several of them related to the same object (DOI), so a non-max-suppression (NMS) algorithm is applied to filter overlapping bounding boxes related to the same object, selecting the one with highest confidence. Two bounding boxes are considered overlapped over the same object when their intersection-over-union (IoU) is higher than a threshold. Considering the dataset pattern used for this work, different objects (DOI) have no, or little bounding boxes overlap, thus a standard threshold of 0.5 for IoU is sufficient for this application.

During training on both model families, the NMS uses a threshold of 0.5 for the IoU and a confidence threshold near zero, to consider all detections for a particular object. For evaluation, different confidence thresholds were tested, using the same IoU threshold, to find the optimal value.

Considering all medical devices used to build the seven-digit dataset, none of them present more than 7 digits at the same time, for instance the MicroLife device family. For this reason, after the NMS algorithm, only the top 7 bounding boxes with highest scores are outputted by the model.

*G. Evaluation Metrics*

After training, the test subset is used for model performance evaluation. This evaluation is divided in two parts. First the MS-COCO challenge metrics [27] are presented and later the detection and classification performance are assessed.

The primary metric used in the COCO challenge is the average precision, defined by Equation 1.

$$AP = \frac{\sum P_c}{N} \quad (1)$$

where $P_c$ is the precision for a particular class, $N$ is the number of classes and $AP$ is the average precision. This metric is calculated for the IoU post-processing threshold of 0.5 ($AP^{0.5}$), the threshold of 0.75 ($AP^{0.75}$) and the mean of all AP's for ten different thresholds between 0.5 and 0.95, the so called mean Average Precision (mAP), which is considered the primary COCO challenge metric.

The same equation is used replacing the precision by the recall for each class, to calculate the Average Recall (AR), using the same IoU thresholds used in mAP. Instead of calculating AR for specific IoU thresholds (0.5 and 0.75), it is calculated for three different maximum number of detections per image: 1 ($AR^1$), 10 ($AR^{10}$) and 100 ($AR^{100}$). For the seven-segment detector, the last ($AR^{100}$) is not necessary, as the maximum number of boxes output is seven, thus this metric is always identical to $AR^{10}$.

In addition to the COCO challenge metrics, the detection and classification performance are assessed. For the detection performance analysis, it is used a fixed value for IoU threshold of 0.5, and three situations are evaluated. The first is when a detected bounding box is close to the ground truth, meaning a true positive result. The second relates to a bounding box proposed by the model where no DOI is present, which is consider a false positive. And the last happens when the model does not output a bounding box for an actual DOI, resulting in a false negative. True negatives are not considered, as the hole background image is considered negative region. These three evaluations are used to calculate the overall precision, recall and F1-score.

Considering only the true positive bounding boxes, the classification performance is evaluated with the confusion matrix plotting, to assess the number of misclassified digits and whether the models struggle to classify any specific digit.

V. RESULTS

In this section, the evaluation metrics are presented using the inference results for all trained models using the test subset (20% of the dataset). First it is presented the MS-COCO challenge metrics, followed by the detection and classification performances.

*A. MS-COCO Challenge Metrics*

Table 2 summarizes the results for all trained models.

TABLE II
MS-COCO CHALLENGE METRICS FOR OBJECT DETECTION MODEL ARCHITECTURES PERFORMANCE.

| Model / Metrics | mAP | $AP^{0.5}$ | $AP^{0.75}$ | $AR^1$ | $AR^{10}$ |
|---|---|---|---|---|---|
| *EfficientDet-0* | 0.809 | 0.989 | 0.961 | 0.672 | 0.840 |
| *EfficientDet-1* | 0.817 | 0.982 | 0.963 | 0.674 | 0.845 |
| *EfficientDet-2* | 0.785 | 0.982 | 0.940 | 0.656 | 0.820 |
| *EfficientDet-lite0* | 0.787 | 0.989 | 0.945 | 0.673 | 0.814 |
| *EfficientDet-lite1* | **0.834** | 0.995 | **0.975** | 0.711 | **0.863** |
| *EfficientDet-lite2* | 0.832 | **0.997** | 0.972 | **0.714** | 0.862 |

Comparing all the results, it is possible to see that models EfficientDet-lite1/lite2 show the best performances for all metrics. Additionally, as stated before, different objects (DOI) have no or little bounding box overlap, so $AP^{0.5}$ metric is more important for the precision analysis, also the maximum number of objects in the same image is 7, thus the most important metrics for recall analysis is $AR^{10}$. For both metrics, the EfficientDet-lite1/lite2 have similar performances.

*B. Detection*

The model's output depends on the IoU and the confidence thresholds. Lower values for the confidence threshold would lead to more low-score bounding boxes, increasing the false positive incidence, while higher values would filter too many boxes, increasing the false negative occurrences. Considering the fixed value for the IoU, different values of confidence thresholds were assessed. For each model architecture, the evaluation metrics were assessed using score thresholds from 0.05 to 0.95 in intervals of 0.05. Figure 5 shows the precision, recall and F1-score by these confidence thresholds plots for each model architecture.

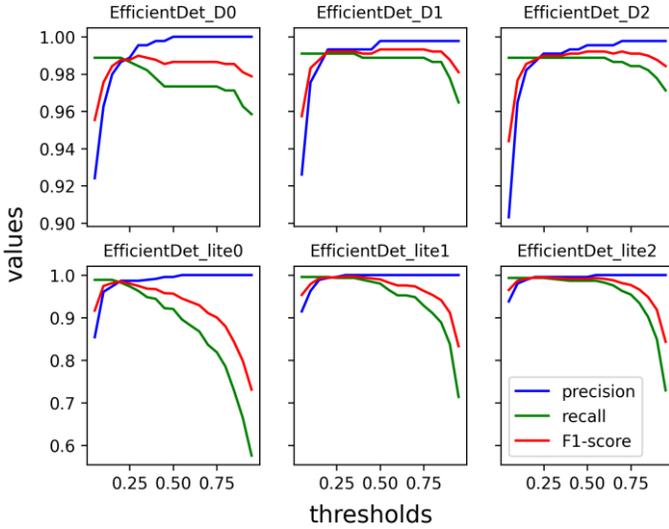

Fig. 5. Precision, recall and F1-score plots by score thresholds applied on detection post-processing stage.

For each model, the optimal score threshold is chosen by the highest F1-score. Running the inference using all 96 images on the test set (438 digits), the evaluation results are summarized in Table 3.

TABLE III
OPTIMAL SCORE THRESHOLD FOR EACH OBJECT DETECTION MODEL ARCHITECTURES AND THEIR PRECISION, RECALL AND F1-SCORE RESULTS.

| Model / Metrics | threshold | precision | recall | F1-score |
|---|---|---|---|---|
| *EfficientDet-D0* | 0.3 | 0.995 | 0.984 | 0.990 |
| *EfficientDet-D1* | 0.5 | 0.998 | 0.989 | 0.993 |
| *EfficientDet-D2* | 0.5 | 0.995 | 0.989 | 0.992 |
| *EfficientDet-lite0* | 0.2 | 0.986 | 0.982 | 0.984 |
| *EfficientDet-lite1* | 0.3 | **1.000** | **0.993** | **0.997** |
| *EfficientDet-lite2* | 0.2 | 0.995 | **0.993** | 0.994 |

EfficientDet-lite1 has the best overall performance, with no false positives and only 3 non detected digits (false negatives) out of 438. In addition, EfficientDet-lite models are usually built specially for mobile applications with smaller size and faster inference, which makes the EfficientDet-lite1 model suitable for cellular phone and tablet apps.

*C. Classification*

As explained in the previous section, only the true positive detected boxes are used for the classification evaluation. For this procedure, it is used the optimal score threshold found before to plot the confusion matrix for each model, as shown in Figure 6.

Again, EfficientDet-lite1, together with EfficientDet-D2 and EfficientDet-lite2 models, show the best performance, with no misclassification, while EfficientDet-D0 and D1 show only 1 misclassification. Even the worst classifier, with 6 wrong outputs out of 424 (EfficientDet-lite0), presents a high accuracy (98.6%).

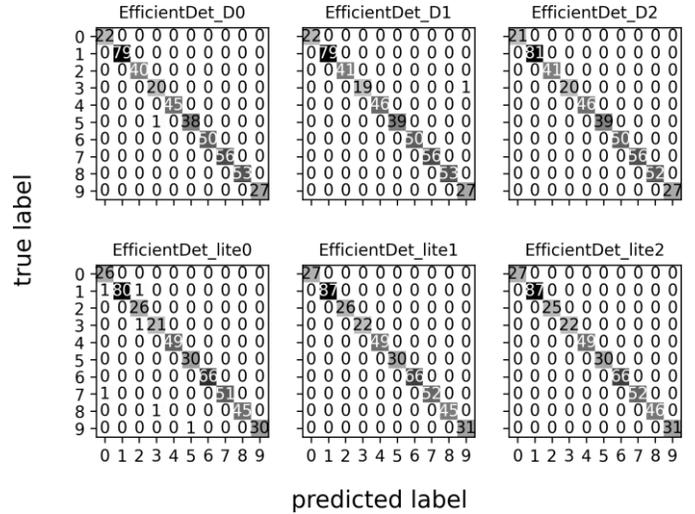

Fig. 6. Confusion matrix for classification performance evaluation. Each cell in the matrices represent the number of digits labelled as shown in the vertical axis (true label) that were classified as the labels in the horizontal axis (predicted label). The diagonal cells represent the number of digits classified correctly.

According to the Tensorflow official webpage [31], the EfficientDet-lite1 has a latency of only 49 milliseconds, on a 4-core CPU smartphone. Added to its above cited detection and classification performances, and its small file size (5.8 MB), EfficientDet-lite1 is suitable for mobile applications.

VI. CONCLUSION

It was proposed in this paper an end-to-end computer model for automated detection of seven-segment characters of medical device displays recorded by mobile cameras. The presented framework is based on state-of-the-art deep-learning-based object detection models for mobile applications. This approach requires minor image pre-processing (only image re-scale) and simple post-processing steps (non-max suppression algorithm).

All models tested show high detection and classification performances, compared to other seven-segment OCR algorithms. The best model, EfficientDet-lite1, presented almost perfect detection metrics, with precision = 1.0 and recall = 0.993, and no classification error. Its performance added to its file size (around 6 MB), makes this model suitable for mobile applications such as smartphones and tablets.

Future work can include the inference of proposed models in real-time image acquisition, using mobile videos as input images; the expansion of the dataset for more medical devices, especially those with different LCD colors, for example oximeters with red displays.

DISCLOSURE STATEMENT

The author reports no conflict of interests.

FUNDING

Author received no specific grant from any funding agency in the public, commercial, or not-for-profit sectors.